\documentclass[conference]{IEEEtran}
\IEEEoverridecommandlockouts
\usepackage{cite}
\usepackage{amsmath,amssymb,amsfonts}
\usepackage{algorithmic}
\usepackage{graphicx}
\usepackage{textcomp}
\usepackage{xcolor}

\usepackage{dsfont}
\usepackage{graphicx}
\usepackage{subcaption}
\usepackage{multirow}
\usepackage{adjustbox}
\usepackage{placeins}
\usepackage{float}
\usepackage{float}
\restylefloat{table}
\usepackage{url}

\def\BibTeX{{\rm B\kern-.05em{\sc i\kern-.025em b}\kern-.08em
    T\kern-.1667em\lower.7ex\hbox{E}\kern-.125emX}}
\begin{document}

\title{GaitSADA: Self-Aligned Domain Adaptation for mmWave Gait Recognition\\
}

\author{
\IEEEauthorblockN{
Ekkasit Pinyoanuntapong\IEEEauthorrefmark{1},
Ayman Ali\IEEEauthorrefmark{1},
Kalvik Jakkala\IEEEauthorrefmark{1},
Pu Wang\IEEEauthorrefmark{1},
Minwoo Lee\IEEEauthorrefmark{1}, \\
Qucheng Peng\IEEEauthorrefmark{2},
Chen Chen\IEEEauthorrefmark{2},
Zhi Sun\IEEEauthorrefmark{3},
\IEEEauthorblockA{\IEEEauthorrefmark{1}University of North Carolina at Charlotte\\
epinyoan@uncc.edu, aali26@uncc.edu, kjakkala@uncc.edu, pwang13@uncc.edu, minwoo.lee@uncc.edu}
\IEEEauthorblockA{\IEEEauthorrefmark{2}University of Central Florida\\
qucheng.peng@knights.ucf.edu, chen.chen@crcv.ucf.edu}
\IEEEauthorblockA{\IEEEauthorrefmark{3}Tsinghua University\\
qzhisun@tsinghua.edu.cn}
}}

\maketitle

\begin{abstract}
mmWave radar-based gait recognition is a novel user identification method that captures human gait biometrics from mmWave radar return signals. This technology offers privacy protection and is resilient to weather and lighting conditions. However, its generalization performance is yet unknown and limits its practical deployment. To address this problem, in this paper, a non-synthetic dataset is collected and analyzed to reveal the presence of spatial and temporal domain shifts in mmWave gait biometric data, which significantly impacts identification accuracy. To mitigate this issue, a novel self-aligned domain adaptation method called GaitSADA is proposed. GaitSADA improves system generalization performance by using a two-stage semi-supervised model training approach. The first stage employs semi-supervised contrastive learning to learn a compact gait representation from both source and target domain data, aligning source-target domain distributions implicitly. The second stage uses semi-supervised consistency training with centroid alignment to further close source-target domain gap by pseudo-labelling the target-domain samples, clustering together the samples belonging to the same class but from different domains, and pushing the class centroid close to the weight vector of each class. Experiments show that GaitSADA outperforms representative domain adaptation methods with an improvement ranging from 15.41\% to 26.32\% on average accuracy in low data regimes. Code and dataset will be available at 
\url{https://exitudio.github.io/GaitSADA}
\end{abstract}


\section{Introduction}
User identification is an essential component for many Internet of Things (IoT) applications, e.g. personalized environmental control, security management, and access control for automatic doors. Short-distance biometrics such as fingerprints, face, iris, palm, and finger vein patterns typically demand active user participation within a certain proximity. In contrast, gait, i.e., human walking style, can be exploited for user identification from a distance without the subject's participation or interference. Moreover, it is shown in existing study \cite{spoof} that human gait is very hard to spoof or mimic because our own gait works against us when we try to imitate someone elses gait. 

The majority of gait recognition techniques are vision-based, which directly perform identification tasks on video data. Appearance-based and model-based approaches are two general approaches for vision-based gait recognition. Appearance-based approaches \cite{GaitPart}\cite{GaitSet}\cite{GaitGL} utilize silhouettes from a video sequence obtained from background subtraction. Although background subtraction is accurate in a lab context, it is difficult to use in a cluttered and dynamic real-world environment. Model-based approaches \cite{GaitMixer}\cite{GaitGraph} try to tackle this issue by employing the existing pose estimators to improve robustness to environmental variation. However, all vision-based approaches require access to RGB visual images which are prone to performance loss from variations of ambient light settings and the social restriction from privacy concerns.

Radar-based gait recognition is an emerging user identification solution that exploits radar return signals to capture human gait biometric \cite{WiWho,WiFiU,NeuralWave,janakaraj2019star,Radar-ID,MU-ID,Gait-based_mmWave}. Unlike camera surveillance systems, radar-based gait recognition systems have several unique advantages. First, they are privacy-preserving as they do not rely on images of human subjects. Next, they can operate in adverse weather and lighting conditions such as heavy fog, smoke, rain, and zero-light conditions. Finally, they can see through some opaque objects depending on its material and the frequency of the radar \cite{wall1,wall2}. Radar-based methods operate by emitting modulated electromagnetic signals, which are scattered and returned back by the subjects in the signal propagation path.  The radar return signals capture the micro-Doppler (mD) signature, which can be represented as radar spectrogram. mD signature or radar spectrogram  documents the time-dependent Doppler frequency shifts, which are caused by the small-scale vibrations or rotations of human body parts during walking.  Each person has his/her unique walking style, i.e., gait, which leads to the unique mD signature that can be exploited for gait recognition. 

Existing radar-based gait recognition systems generally operate either in sub-6GHz WiFi bands \cite{WiWho,WiFiU,NeuralWave} or the emerging mmWave bands (30 to 300 GHz) \cite{janakaraj2019star,Radar-ID,MU-ID,Gait-based_mmWave}. Compared with sub-6GHz radar systems, mmWave radar can promise much fine-grained motion capture with significantly enhanced velocity and location resolutions. This feature is enabled by orderly shorter wavelength and wider bandwidth of mmWave radar signals, compared with the sub-6GHz counterparts. Besides promising motion capture capacities, mmWave radar system only needs a single radar device that integrates both radar transmitter and receiver, while sub-6GHz radar system is generally implemented by a pair of WiFi devices to act as distributed radar transmitter and receiver, which have to run modified WiFi firmware to extract mD signatures from CSI signals. 

The superior gait recognition performance of mmWave radar systems has been demonstrated in our previous research \cite{janakaraj2019star} and other related work \cite{Radar-ID,MU-ID,Gait-based_mmWave}. However, the spatial and temporal generalization performance of this technology is still unknown. We found that deep neural networks (DNNs) trained on radar-based gait data suffer from substantial performance degradation with the progression of time and in new locations. Existing radar-based gait recognition systems proposed in the literature, test their performance on a dataset collected alongside the training data and fail to uncover the temporal degradation. Furthermore, they seldom address the spatial generalization issues from introducing new environments. The root cause of this generalization issue is domain shift \cite{moreno2012unifying}. Simply put, it occurs when the testing data distribution differs from that of the training data, i.e., the i.i.d. assumption is violated. Domain shift problem can be further exacerbated by the harmful environment reflections induced by mmWave signals.  These environment reflections include static ones, which directly bounce from stationary obstructions e.g. walls, desks, and chairs, and dynamic ones,  which indirectly bounce from other stationary obstructions and then bounce from human bodies. The harmful environment reflections carry delayed and distorted mD signature information, which cannot be completely removed. This leads to environment-induced temporal domain shift (TDS) and spatial domain shift (SDS). Moreover, human gait, although unique to each individual, could contain many variations, potentially as a consequence of a subject's mood and clothing changes, further aggravate TDS and SDS.



In this paper, we tackle the challenge of spatial-temporal domain drift problem in mmWave radar-based gait recognition. To address this issue, we propose a novel domain adaptation method, GaitSADA. GaitSADA is based on simple distribution alignment components and consists of a two-stage approach: pretraining and fine-tuning. In the pretraining stage, the model learns salient gait representations and noise invariance from both source and target domains using semi-supervised contrastive learning. In the fine-tuning stage, the model generalizes the learned gait representation to the target domain by leveraging pseudo-labels, semi-supervised consistency training, and centroid alignment. Experiments demonstrate that GaitSADA outperforms classic adversarial domain adaptation methods and achieves significant accuracy improvements in low data regimes and challenging domain drift conditions.  Our contributions can be summarized as follows:

\begin{itemize}
 \item We curate a non-synthetic dataset consisting of mmWave radar-based gait biometric data. This dataset, for the first time, allows one to study and improve the spatio-temporal generalization performance of a radar-based biometric identification system.
 \item We perform extensive research on the SDS and TDS from four different location environments on different days, experimenting with five state-of-the-art (SOTA) domain adaptation methods to mitigate these drifts and show the issue of accuracy drop in low data regimes of SOTA domain adaptation methods.
 \item We propose a novel domain adaptation method for mmWave gait recognition system, self-aligned domain adaptation for mmWave gait recognition (GaitSADA), by jointly exploiting semi-supervised contrastive learning, semi-supervised consistency training, centroid alignment and deliberately designed radar spectrogram augmentations. GaitSADA outperforms existing SOTA methods in 1-day, 2-day, and 3-day data by 15.41\%, 5.83\%, and 2.63\% respectively on average of domain drift cases.
\end{itemize}

\section{Related Work}
Unsupervised domain adaptation (UDA), semi-supervised learning,  and self-supervised learning are common strategies to reduce the dependency on expensive manually annotated data. UDA aims to mitigate the domain gap by transferring knowledge learned from label data in source domain and transfer to unlabeled data in target domain, while semi-supervised learning employs a few manually annotated samples and generalizes to large unlabeled samples without domain shift assumption. Self-supervised utilizes the unlabeled data to pre-train the model to learn the general semantics of the data.


Unsupervised domain adaptation can tackle the domain drift and the lack of labeled data problem by minimizing the discrepancy between source and target distributions without using expensive labeled data from the target domain. Many UDA techniques, inspired by Generative Adversarial Networks (GAN), apply an adversarial loss to confuse the domain discriminator network that tries to recognize the domain difference. ADDA \cite{ADDA} exploits  two networks for source and target domains, respectively and forces the output representation to be in the same distribution. Domain-adversarial neural network (DANN) trains domain discriminator along with the feature extractor that confuses the domain discriminator by integrating a gradient reversal layer (GRL) \cite{GRL}. Although the overall distribution is aligned, the class-wise discrepancy can still cause performance drop in the classification tasks. Conditional Adversarial Domain Adaptation (CDAN) \cite{CDAN} takes this into account by implementing multilinear conditioning of the feature representation and the class-wise output prediction. 


Semi-supervised methods decrease the dependency on labeled data by combining normally smaller sets of labeled data with the vast amounts of unlabeled data that are available in many use cases. Since the classification loss is unknown for unlabeled data, the state-of-the-art (SOTA) semi-supervised Learning solutions, such as UDA \cite{UDA}, FixMatch \cite{FixMatch}, MixMatch \cite{MixMatch}, and RemixMatch \cite{ReMixMatch}  mainly rely on consistency training, which learns target unlabeled data $u_i$ by generating pseudo-labels from models' predictions on weakly-augmented data  and filtering only high confidence data by selecting the prediction that has  confidence more than the threshold. The cross-entropy $H$ is applied against the pseudo labels and the model’s prediction of the highly-augmented versions of the same unlabeled data. 

Similar to semi-supervised learning, self-supervised Learning utilizes large unlabeled data to improve performance from label data trained in a supervised learning fashion. However, self-supervised learning normally pre-trains unlabeled data to learn general data semantics before transferring pre-trained weights to further learn label data in a supervised learning fashion in the next stage. Many research \cite{SimCRL}, \cite{BYOL}, \cite{infoNCE}, \cite{MoCo} enforce contrastive loss to learn the structure of the data without labels by imposing different augmentation to the same data and learning the similarity between the features of same data (positive samples). The challenge is that such similarity loss functions can trigger a dimensional collapse in which the model outputs the exact same embedding for all input. While MoCo \cite{MoCo} and BYOL \cite{BYOL} incorporate momentum encoder and stop gradient to prevent the collapse, SimCRL \cite{SimCRL} and CPC \cite{infoNCE} employ negative samples to discriminate the features of different data. SimCLR \cite{SimCRL} proposed NT-Xent (the normalized temperature-scaled cross-entropy loss) which simply maximizes the similarity between the feature of the same data (positive samples) and minimizes the different ones (negative samples).

\section{mmWave gait biometric Dataset} \label{sec:system_design}
The dataset is created by exploiting the mmWave radar-based gait recognition system, STAR, developed in our previous work \cite{janakaraj2019star}. In this section, the preliminaries of mmWave radar sensing are first presented, the data collection and gait recognition procedure using STAR system is then introduced, and finally spatial-temporal domain drift issues are revealed.

\begin{figure}[t]
	\centering	\includegraphics[width=1.0\linewidth]{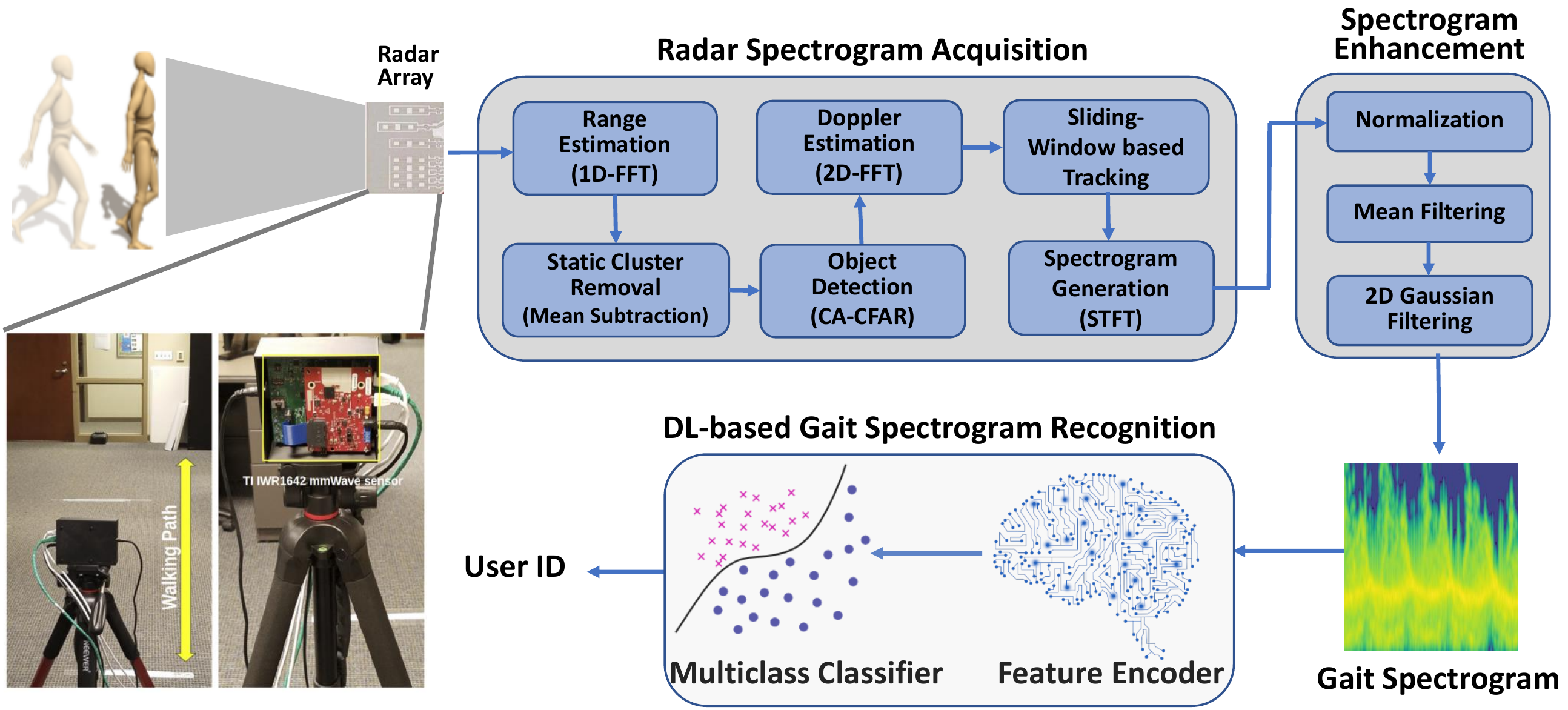} 
	\caption{Overview of STAR system}
	\label{fig:STAR}
\end{figure}

\subsection{STAR System Overview}
STAR is the first one in the literature to exploit deep learning for mmWave radar-based gait recognition. As shown in Fig. \ref{fig:STAR}, STAR consists of three components: radar spectrogram acquisition, spectrogram enhancement and deep learning(DL)-based gait spectrogram recognition. The spectrogram acquisition module tracks and captures the radar signal returns that directly bounce off from human body, while filtering out undesired environment-induced radar returns. This is achieved by applying a sequence of radar widely-adopted radar signal processing algorithms including one-dimensional Fourier transform (ID-FFT) for range estimation, the mean subtraction for static reflection removal, CA-CFAR for robust target detection against a background of noise, clutter and interference, Doppler estimation via 2D-FFT to estimate target velocity, and sliding-window based tracking method to predict the target future state (e.g., location and velocity). The target tracking module acts a moving spatial passband filter so that only the radar returns from the target can be received. Performing short-time Fourier transform (STFT) on received radar returns generates the raw radar
spectrogram, whose quality is further improved by exploiting the spectrogram enhancement module that consists of a sequence of frequency-domain noise removal processes \cite{janakaraj2019star}. The enhanced spectrograms are then encoded by a CNN-based feature encoder that maps the high-dimension spectrogram data into the compact low-dimension feature vector. Finally, the compact feature vectors are used to train a multiclass classifier that predicts the corresponding user identities. 
\subsection{Data Collection Setup}
\begin{figure}[t] 
\centerline{\includegraphics[width=.5\textwidth]{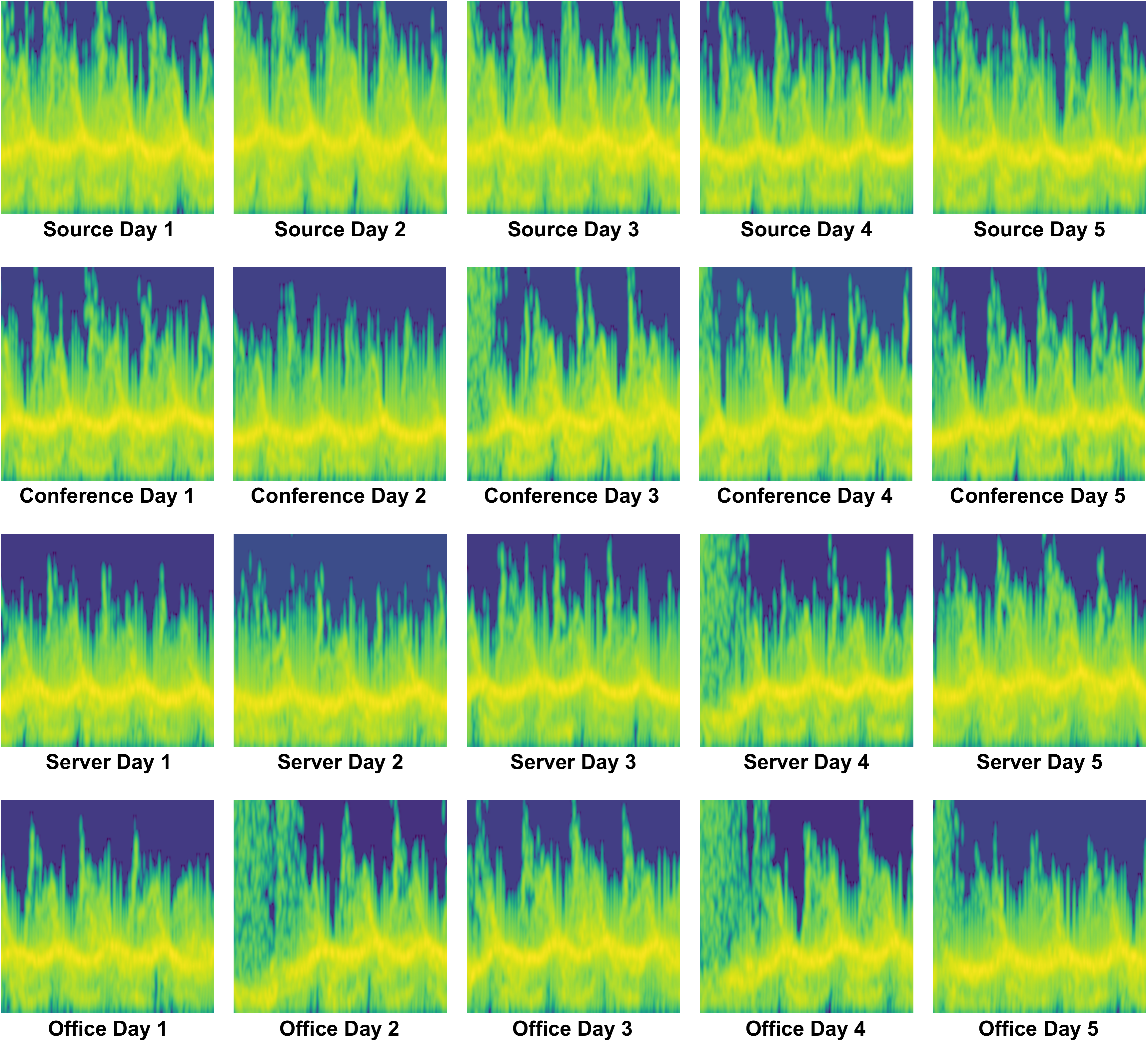}}
\caption{Gait spectrograms of the same person for 5 days at different locations, source (laboratory), conference room, server room, and office.}
\label{fig:spec_sample}
\end{figure}

 We collected gait data from 10 volunteers between the ages of 18-35. Each subject's data was collected in four different locations. The source location was a research space with cubicles, and the other three areas consisted of a server, conference, and an office room. By maintaining four distinct locations, we introduce SDS. In the source location, data was collected on 10 different days for each subject. 5 separate days of data was acquired for each of the three other locations, which are used as target domains. A participant can either walk towards the radar or walk away from the radar, each of which is counted as one walking instance and generates one spectrogram data sample. The data collection was limited to 100 data samples per person in the source location and 50 data samples per person in each of the target locations on any given day. The gait spectrogram samples of a person are shown in Fig. \ref{fig:spec_sample}.

\subsection{Establishing the Presence of TDS and SDS}
We study the presence of TDS and SDS issues by training gait recognition model using the labelled data from source domain and test the model performance across different target domains. In particular, the feature encoder is a modified version of ResNet-50 \cite{ResNet}, as described in section \ref{subsec:Model}, to extract features from mmWave biometric data samples. The loss function is a cross-entropy loss for multiclass classification. To establish the presence of SDS, we train the model on the data from 1 to 3 days of source location and test the model on the data from other three unseen locations. SDS is evident from the performance degradation in all target domains as shown in table \ref{tab:supervised_learning}. To reveal the existence of TDS, we train the model on the 
data from 1 to 3 days of the source location and 
 test the model on the data from the remaining 
days of the source location. TDS can be observed due to the degraded testing accuracy.


	

\begin{table}
\centering
\caption{The result of supervised learning trained on source data and tested on different days of data (i.e., temporal) and other different environments (i.e., server, conference, office). The accuracy drop reveals the domain discrepancy when the experiments are conducted on different days and environments. The results report in \% of accuracy.}
\label{tab:supervised_learning}
\begin{adjustbox}{width=1.0\linewidth,center}
\begin{tabular}{|c|c|c|c|c|} 
\hline
\begin{tabular}[c]{@{}c@{}}\textbf{\# of training days }\\\textbf{(Source)}\end{tabular} & \textbf{Temporal} & \textbf{Server~} & \textbf{Conference~ ~} & \textbf{Office}  \\ 
\hline
1-day                                                                                    & 76.05             & 34.93            & 47.00                  & 33.39            \\ 
\hline
2-days                                                                                   & 88.46             & 53.05            & 71.00                  & 49.92            \\ 
\hline
3-days                                                                                   & 95.25             & 71.77            & 85.20                  & 67.61            \\
\hline
\end{tabular}
\end{adjustbox}
\end{table}

\section{Method: GaitSADA}
\label{method:our}
\subsection{Method Overview}
\begin{figure*}[t!]
    \centering
    \begin{subfigure}[b]{1.0\textwidth}
        \centering
        \includegraphics[width=0.8\linewidth]{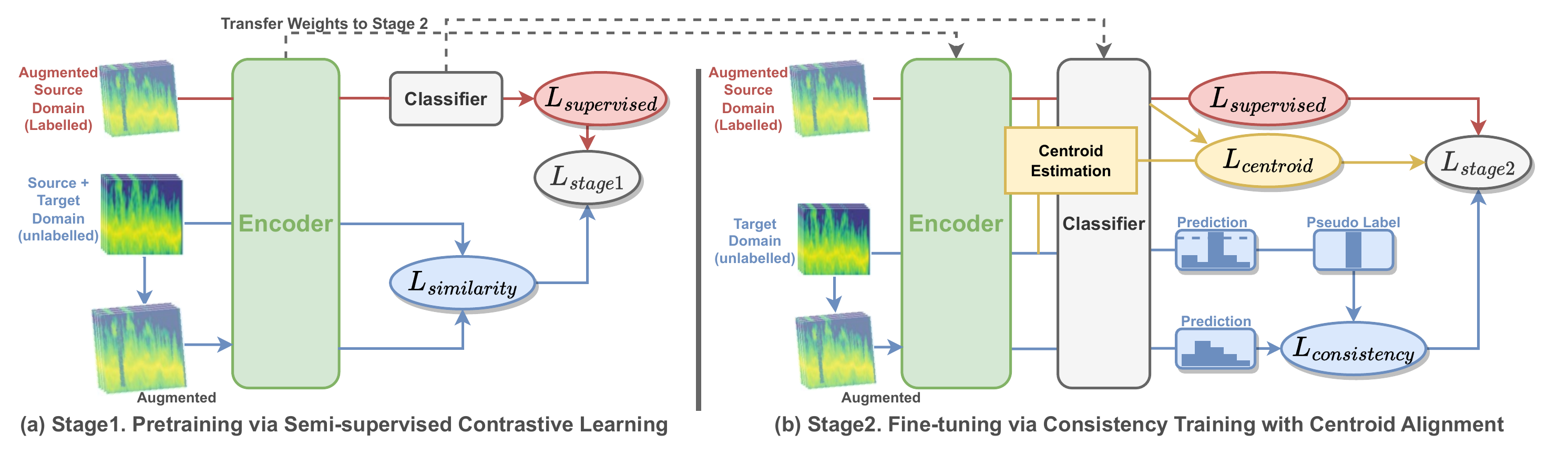} 
    \end{subfigure}

    \caption{\textbf{Overall Architecture of GaitSADA.} \textbf{(a) Stage 1.} Pretraining via semi-supervised contrastive learning to learn a compact gait representation from both source and domain data, which also implicitly mitigates the source-domain distribution shift. \textbf{(b) Stage 2.} fine-tuning via semi-supervised consistency training with centroid alignment. The fine-tuning stage aims to further improve model generalization performance by pseudo-labelling the target-domain samples, clustering together the samples belonging to the same class but from different domains, and pushing the class centroid close to the weight vector of each class. The encoders are sharing the weights.}
    \label{fig:overall_architecture}
\end{figure*}

Our novel method, self-aligned domain adaptation for mmWave gait recognition (GaitSADA), jointly exploits self- and semi-supervised learning techniques to realize the unsupervised domain adaptation by promoting intra-class compactness and inter-class distance while generalizing knowledge from label data to unlabeled data using pseudo-label. In particular, we leverage the unlaballed data from the target domains in two training stages: pretaining and fine-tuning as illustrated in Fig.~\ref{fig:overall_architecture}.

During pretraining stage, we propose a semi-supervised contrastive  learning paradigm to learn the salient gait representation from a mixed dataset that consists of data from both source and and target domains. Our semi-supervised contrastive  learning can explicitly avoid collapse problem without requiring the sophisticated techniques adopted by classic self-supervised (unsupervised) contrastive  learning paradigms, e.g., negative samples, stop gradients, and moment encoders. This is achieved by maximizing the agreement between representations produced by encoders fed with different augmentations of the same spectrogam in the mixed dataset, while enforcing the encoder fed with labelled data in source domain to produce discriminative representations with correct class predictions. 

After the pretrained stage, the feature encoder implicitly aligns the data distributions of source domain and target domain in the feature space. As a result, the representations produced by such encoder are discriminative enough to provide relatively high-confidence pseudo labels for unlabelled data in the target domains, which is essential for the semi-supervised consistency training in the fine-tuning stage that can effectively minimize the embedding vectors of the target domain data from the same class to be close to each other. However, classic semi-supervised consistency training only shows the best performance when the label and unlabelled data come from the same distribution. To address this issue, we propose the centroid alignment scheme that can be integrated with semi-supervised consistency training to mitigate source-target domain gap.

\subsection{\textbf{Pretraining via Semi-supervised Contrastive  Learning}}
\label{subsection:AM_self_alingment}

\noindent\textbf{Supervised Learning with Additive Margin Softmax} 
The main goal of this component is to promote intra-class compactness and inter-class separability. As the regular softmax loss is effective only in maximizing the inter-class difference, it is ineffective at minimizing the intra-class variation. Therefore, we adopt Additive Margin Softmax loss function \cite{AdditiveMargin} to bring the classification boundary closer to the weight vector of each class by adding a margin $m$ to the boundary as

\begin{equation}\label{eq:additive_margin}
    L_{supervised} = -\frac{1}{B}\sum_{i=1}^{B}log\frac{e^{s\cdot(cos\theta_{y_i})-m  }}{  e^{s\cdot(cos\theta_{y_i})-m}+\sum_{j=1,j\neq y_i}^{C} e^{s\cdot cos\theta_j} },
\end{equation}
where $cos\theta_{y_i} =\frac{W^T_{y_i} f_i}{ \left \| W_{y_i} \right \| \left \| f_i \right \|}$ represents cosine similarity of the inner product between the normalized weight $W_{y_i}$ and the normalized feature $f_i$ of class $i$. Additional parameter $s$ is a scale parameter to control the temperature. \\

\noindent\textbf{Positive Unlabelled Contrastive  Learning.}
The objective of the second component is to learn noise invariance by training the model to generate the same features regardless of the artificial noise that is added to the data without relying on label data. We carefully design multiple augmentations to simulate the radio wave noise such as frequency noise, temporal noise, different resolutions, and white noise as described in Section~\ref{subsection:augmentation}. The model learns noise invariance by minimizing cosine similarity between features from raw data $f$ and its augmented version $\hat{f}$ with the loss

\begin{equation}
L_{similarity} =\frac{1}{B}\sum_{i=1}^{B}\frac{ f_i^T \hat{f_i}}{ \left \| f_i \right \| \left \| \hat{f_i} \right \|}.
\end{equation}
This loss is inspired by NT-Xent lose from SimCLR \cite{SimCRL} which is the contrastive self-supervised learning method that minimizes cosine similarity of positive samples (i.e., different augmentation of the same data) and maximizes the cosine similarity of negative samples (i.e., different data). The objective of negative samples in NT-Xent is to prevent dimensional collapse. However, since our self alignment loss is trained jointly with supervised learning loss, this can explicitly avoid the collapse issue without relying on negative samples. This is because if the encoder produces constant embedding vectors for different spectrograms (i.e., dimensional collapse occurs) via self-supervised learning branch, this will prevent supervised learning from correctly predicting the classes of the labelled spectrograms. As a result, we are able to retain only the positive similarity and eliminate the negative marginal fraction. The final loss function in the first stage combines two loss functions to encourage the intra-class concentration and inter-class distance while being invariant to perturbation from noise by utilizing unlabeled data from target domain:
\begin{equation}
L_{stage1} = L_{supervised}+L_{similarity}.
\end{equation}

\subsection{\textbf{Fine-tuning via Consistency Training with Centroid Alignment}}

In the first stage, the model  has learned the well-aligned class-wise feature from label data in the source domain and unlabelled data from target domain. The objective of the second stage is to further regulate the feature of the target domain data of the same class to be close to each other. \\ 

\noindent\textbf{Consistency Training.} Unlike the source domain, the labels in the target domain are unknown. Therefore, we cannot directly align target feature by class. To address this problem, we enforce consistency training via pseudo labels instead of the true labels of the target domain. Consistency training utilizes unlabeled data by relying on the assumption that when given perturbed variations of the same input, the model ought to provide similar predictions. The two identical models are used for consistency training. The first branch takes raw input data with no augmentation in order to provide accurate pseudo labels. We compute the model’s predicted class distribution of a given unlabeled spectrogram input: $q_i = p(y_i | u_i)$ as soft pseudo labels. The cross-entropy $H$ is applied against the pseudo labels distribution and the model’s prediction distribution of the augmented unlabeled data $\mathcal{A}(u_i)$. The loss is defined as 
\begin{equation} \label{eq:consistency}
L_{consistency} = \frac{1}{B}\sum_{i=1}^{B} \mathds{1} (\arg \max_{y} (q_i)\geq \tau) H(q_i, \hat{q}_i),
\end{equation}
where $\hat{q}_i=p(y_i | \mathcal{A}(u_i))$ represents the probability distribution of class $y_i$ given augmented input $\mathcal{A}(u_i)$ and $\tau$ is a scalar parameter denoting the threshold to filter only the high confidence prediction. To maintain the same standard of Additive Margin Softmax (Eq.~\eqref{eq:additive_margin}) in the first stage, we normalize weights and features of the classifier which can be referred to as cosine similarity $cos\theta$. The scale $s$ remains the same, but we remove margin $m$ for accurate pseudo labels:
\begin{equation}
p(y_i|\cdot ) = \frac{e^{s\cdot(cos\theta_{y_i})  }}{  \sum_{j=1}^{c} e^{s\cdot cos\theta_j} },
\end{equation}
where $c$ is the number of classes. The predictions are likely to be inaccurate at the beginning of the training. For this, most semi-supervised learning algorithms \cite{MeanTeacher}\cite{TemporalEnsembling}\cite{MixMatch}\cite{ReMixMatch} initialize training with a small weight $\lambda$ for the consistency loss term and increase it over time. However, with the confidence threshold, the pseudo label produces a natural curriculum learning \cite{FixMatch}. As less pseudo-label data are available at first since the low confidence predictions are filtered out. When the model is more accurate, it then provides more predictions with a high degree of confidence. As a result, the threshold automatically increases the number of pseudo-label data without explicitly specifying incremental weight $\lambda$.

\begin{figure}[t] 
\centerline{\includegraphics[width=.3\textwidth]{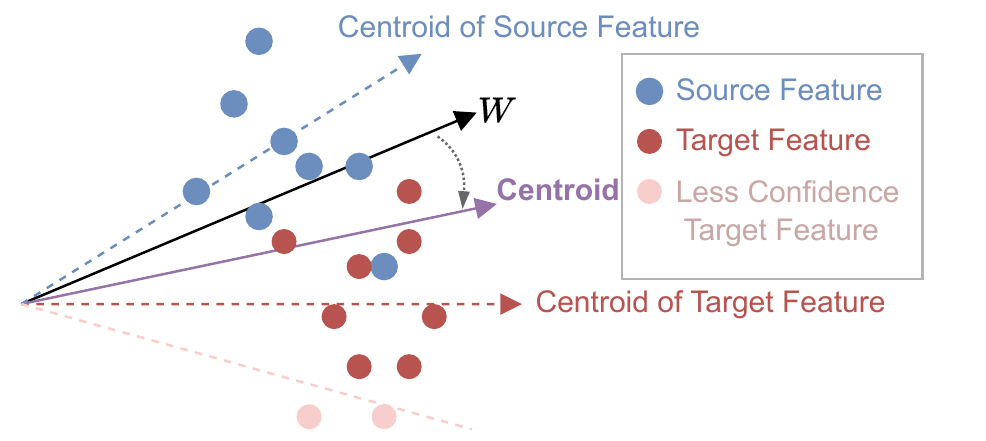}}

\caption{\textbf{Centroid Alignment.} The centroid of source features tends to be closed to weight vector $W$ of the corresponding class following the supervised learning loss. On the other hand, the centroid of target features is likely to be further away due to the domain gap. Centroid Alignment encourages weight vector $W$ to move to a new centroid which is the center of both source and target features that can be a better representation as more  target features are learned from consistency training.}
\label{fig:centroid}
\end{figure}

\noindent\textbf{Centroid Alignment.} While consistency training loss tends to move weights $W$ of the classifier toward target features, supervised learning loss still attempts to bring the weights back to source features. The weights can oscillate between these two groups as a result of the domain gap between the source and target features as shown in Fig.~\ref{fig:centroid}. To mitigate this issue, we impose centroid alignment loss to gradually bring the weights to the center of source and target features by applying Additive Margin Softmax to class-wise weights and class centroids as

\begin{equation}
    L_{centroid} = -\frac{1}{C}\sum_{c=1}^{C}log\frac{e^{s\cdot(cos\theta_{c})-m  }}{  e^{s\cdot(cos\theta_{c})-m}+\sum_{j=1,j\neq i}^{C} e^{s\cdot cos\theta_j}},
\end{equation}
where $cos\theta_{c} =\frac{W^T_{c} Z_c}{ \left \| W_{c} \right \| \left \| Z_c \right \|}$ is the inner product of normalized weights $W_i$ and centroids $Z_i$ of each class. The class-wise centroid is from the Exponential Moving Average (EMA) of both source and target features with hyperparameter $\alpha$ to control the rate of change as

\begin{equation}
Z_{c_i} = \alpha \hat{Z}_{c_i} + (1-\alpha)Z_{c_{i-1}}.
\end{equation}
The current batch centroid $\hat{Z}_{c_i}$ is the average features of source $f_{s_i}$ and target $f_{t_i}$. However, because the true class of the target domain is unknown, we cannot directly assign features to the class centroid. Instead, the hardmax (i.e., argmax) of the model’s prediction is used as a pseudo label of the class of each sample. Also, target features can be noisy at the beginning of the training, similar to consistency loss (Eq.~\eqref{eq:consistency}), only the high confidence labels are used for calculating the centroids to avoid false predictions of pseudo labels. Therefore,

\begin{equation}
\hat{Z}_{c_i} = \frac{\sum_{i=1}^{B} f_{s_i} + \mathds{1} (\arg \max_{y} (q_i)\geq \tau) f_{t_i}}{B+\sum_{i=1}^{B} \mathds{1} (\arg \max_{y} (q_i)\geq \tau)}.
\end{equation}
Lastly, we impose Additive Margin Softmax for source label data as a supervised learning loss along with consistency training and Centroid Alignment loss to prevent dimensional collapse. So the final loss equation is shown as 
\begin{equation}    
L_{stage2}=L_{supervised} + L_{consistency} + \lambda L_{centroid}.
\end{equation}

\section{Experiment Setup}
Texas Instruments (TI) IWR1642EVM boost board interfaced with a DCA1000EVM board to collect the raw mmWave radar signal. The radar system consists of two transmitting antennas, four receiving antennas, and 120° view of the azimuth plane. The radar system supports up to a 4Ghz bandwidth operating on 77 GHz to 81 GHz. To configure FMCW wave parameters such as chirp width, repetition time, and chirp slope from our radar device, we use a Dell Latitude 7480 laptop with TI mmWave studio software as a control system.


Laboratory location data is used as source data and the other location data (i.e. conference room, server room, and office) is referred to as target data. For the spatial domain drift experiment, day 1 to day 3 samples are used as a training set for all source and target locations. The rest of the data is used for testing. For the temporal domain drift experiment, we employ 1 to 3 days of laboratory location as source data and the next consecutive days of laboratory location as target data. For example, considering the temporal 2-day case, the first and the second-day data of laboratory location is a source domain, the third and the fourth-day data is a target domain, and the fifth to the tenth-day data is utilized as a test set.

\subsection{Spectrogram Augmentation} 
\label{subsection:augmentation}

\begin{figure}[!tbp]
  \centering
  \begin{subfigure}[b]{0.15\textwidth}
    \centering
    \includegraphics[width=\linewidth]{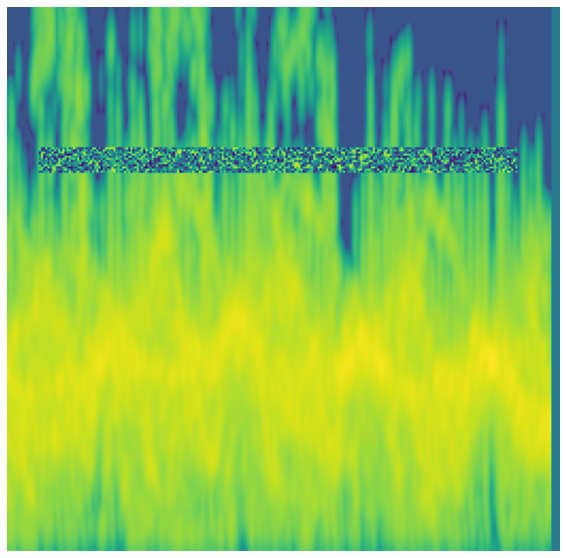}
    \captionsetup{justification=centering}
    \caption{\footnotesize Cutout\\Horizontal \\ \color{white}.}
    \label{fig:aug_horizon}
  \end{subfigure}
  \begin{subfigure}[b]{0.15\textwidth}
    \centering
    \includegraphics[width=\linewidth]{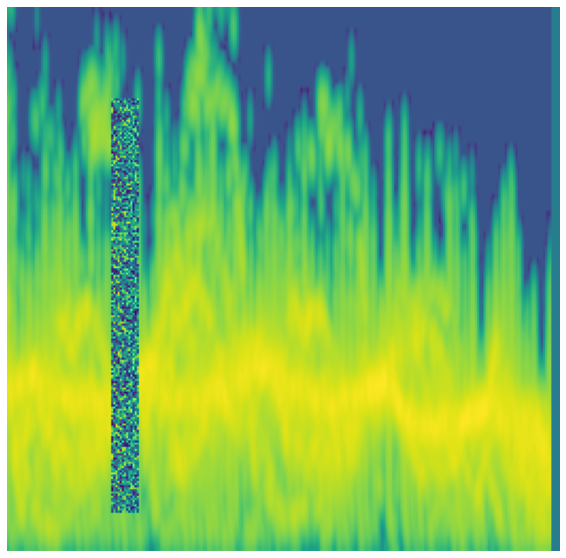}
    \captionsetup{justification=centering}
    \caption{\footnotesize Cutout \\ Vertical \\ \color{white}.}
    \label{fig:aug_vertical}
  \end{subfigure}
  \begin{subfigure}[b]{0.15\textwidth}
    \centering
    \includegraphics[width=\linewidth]{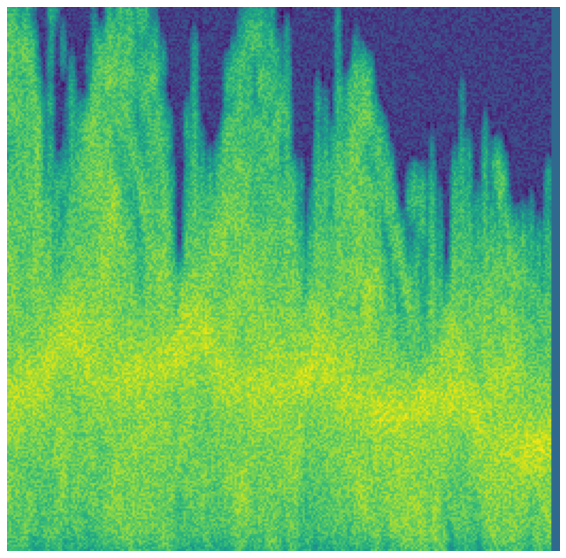}
    \captionsetup{justification=centering}
    \caption{\footnotesize White Noise \\ \color{white}. \\ \color{white}.  }
    \label{fig:aug_white}
  \end{subfigure}

  \begin{subfigure}[b]{0.15\textwidth}
    \centering
    \includegraphics[width=\linewidth]{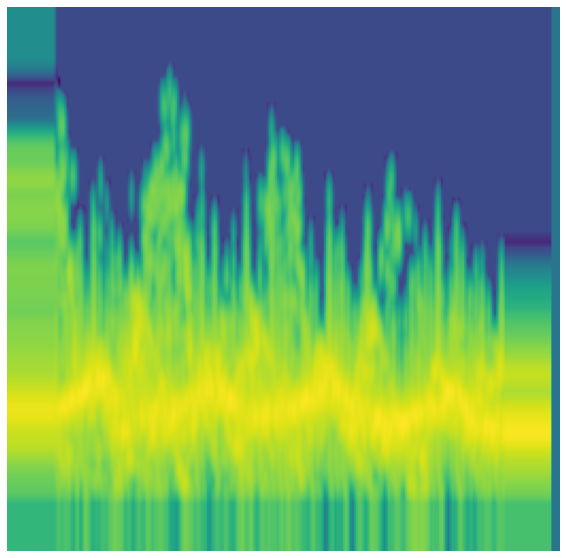}
    \captionsetup{justification=centering}
    \caption{\footnotesize Zoom \\ \color{white}.}
    \label{fig:aug_zoom}
  \end{subfigure}
  \begin{subfigure}[b]{0.15\textwidth}
    \centering
    \includegraphics[width=\linewidth]{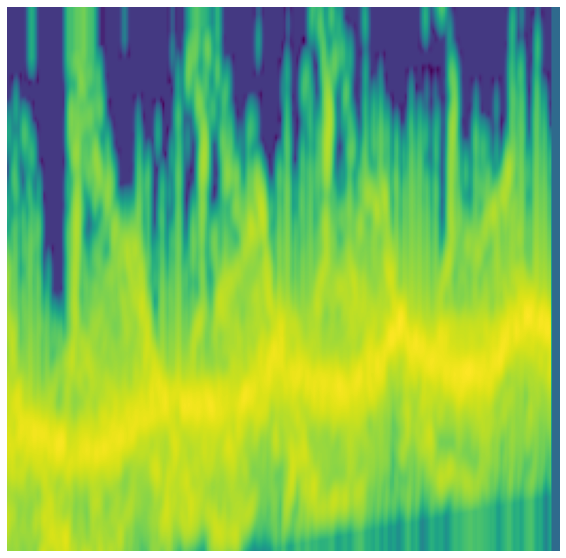}
    \captionsetup{justification=centering}
    \caption{\footnotesize Shear \\ \color{white}.}
    \label{fig:aug_shear}
  \end{subfigure}
  \begin{subfigure}[b]{0.15\textwidth}
    \centering
    \includegraphics[width=\linewidth]{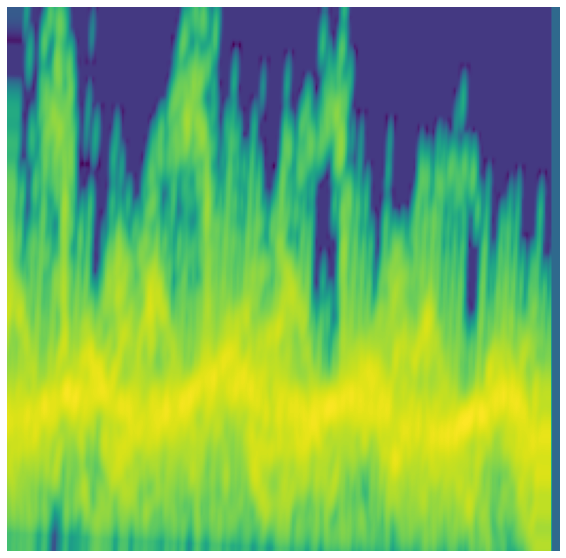}
    \captionsetup{justification=centering}
    \caption{\footnotesize Rotation \\ \color{white}.}
    \label{fig:aug_rot}
  \end{subfigure}
    \caption{Spectrogram augmentations.}
    \label{fig:augmentation}
\end{figure}


\begin{table}[h!]
\caption{Result of training on 1 to 3 days on the data from laboratory location (source domain) while adapting to different 1 to 3 days data of same location (i.e., temporal target domain) and 1 to 3 days of different target locations, (i.e., server, conference, and office) with 5 representative methods, GRL \cite{GRL}, GAN \cite{GAN}, ADDA \cite{ADDA}, CDAN \cite{CDAN}, and FixMatch \cite{FixMatch} along with "Supervised Learning" which train only on source label data without domain adaptation and lastly, our proposed method, GaitSADA. The results report in \% of accuracy.  }

\centering
\begin{adjustbox}{width=1.0\linewidth,center}
\begin{tabular}{c|c|cccc|c} 
\hline
\textbf{\# of training days}    & \textbf{Methods}                & \textbf{Temporal} & \textbf{Server} & \textbf{Conference} & \textbf{Office} & \textbf{Average}  \\ 
\hline
\multirow{7}{*}{\textbf{1-day}} & Supervised                      & 76.05             & 34.93           & 47.00               & 33.39           & 47.84             \\
                                & GAN                             & 79.57             & 38.14           & 66.00               & 32.93           & 54.16             \\
                                & GRL                             & 74.97             & 40.14           & 62.10               & 36.14           & 53.34             \\
                                & ADDA                            & 80.96             & 37.71           & 65.21               & 32.60           & 54.12             \\
                                & CDAN                            & 73.67             & 45.15           & 62.40               & 43.24           & 56.12             \\
                                & FixMatch                        & 94.32             & 44.24           & 69.40               & 49.05           & 64.25             \\ 
\cline{2-7}
                                & \textbf{GaitSADA (our)} & \textbf{96.3}     & \textbf{75.98}  & \textbf{82.6}       & \textbf{63.76}  & \textbf{79.66}    \\ 
\hline
\multirow{7}{*}{\textbf{2-day}} & Supervised                      & 88.46             & 53.05           & 71.00               & 49.92           & 65.61             \\
                                & GAN                             & 92.18             & 59.96           & 77.80               & 59.76           & 72.43             \\
                                & GRL                             & 85.98             & 69.07           & 80.90               & 59.96           & 73.98             \\
                                & ADDA                            & 88.55             & 51.04           & 75.63               & 55.62           & 67.71             \\
                                & CDAN                            & 91.72             & 63.36           & 85.90               & 63.16           & 76.04             \\
                                & FixMatch                        & 95.82             & 75.38           & 90.10               & 82.58           & 85.97             \\ 
\cline{2-7}
                                & \textbf{GaitSADA (our)} & \textbf{97.63}    & \textbf{79.88}  & \textbf{99.1}       & \textbf{90.59}  & \textbf{91.8}     \\ 
\hline
\multirow{7}{*}{\textbf{3-day}} & Supervised                      & 95.25             & 71.77           & 85.20               & 67.61           & 79.96             \\
                                & GAN                             & 96.79             & 78.18           & 94.30               & 85.89           & 88.79             \\
                                & GRL                             & 96.11             & 84.58           & 89.30               & 75.58           & 86.39             \\
                                & ADDA                            & 97.38             & 79.79           & 94.79               & 88.13           & 90.02             \\
                                & CDAN                            & 98.03             & 91.39           & 93.80               & 86.19           & 92.35             \\
                                & FixMatch                        & 99.29             & 93.29           & 98.30               & 88.89           & 94.94             \\ 
\cline{2-7}
                                & \textbf{GaitSADA (our)} & \textbf{99.68}    & \textbf{93.69}  & \textbf{99.2}       & \textbf{97.7}   & \textbf{97.57}    \\
\hline
\end{tabular}
\label{tab:experiment}
\end{adjustbox}
\end{table}
Vertical noise stripe (Fig.~\ref{fig:aug_vertical}) simulates missing information in the random time interval of the spectrogram. Similarly, the horizontal noise stripe (Fig.~\ref{fig:aug_vertical}) represents the missing information in some frequency range. We implement horizontal and vertical noise together by randomly adding two of these noises per data with a probability of 1/3 to be horizontal and 1/3 to be vertical, the rest 1/3 is for no cutout noise. The small side of the stripe is random between 2 to 8 pixels. 
The third augmentation is pixel-wised random numbers adjusted to the whole data uniformly simulating white noise (Fig.~\ref{fig:aug_white}) in a spectrogram. The uniformly random noise from -0.5 to 0.5 is added to the data with a probability of occurrence 2/3. Next is the zoom-in and out effect. Zoom-in will drop some information from the start and the end, also the lowest and the highest frequency of the spectrogram. Zoom effects (Fig.~\ref{fig:aug_zoom}) represent the data in different resolutions. The scale is applied to the data between 0.8 to 1.2 randomly. These four augmentations act as virtual noise to simulate the noisy real-world data thereby improving domain generalization. Not only these meaningful noises, but we also experiment with regular computer vision augmentations and we found that shearing (Fig.~\ref{fig:aug_shear}) and rotation (Fig.~\ref{fig:aug_rot}) help boost performance as well. The shear angle in a counter-clockwise direction in degrees is random from 0 to 5. Meanwhile, 5-degree ranges are randomly applied for rotation. We apply these augmentations to our preprocess pipeline and then apply all augmentation to every spectrogram data with random magnitude. In addition to consistency training, augmentation is used in supervised learning during the first training stage.

\subsection{Model and Training Details}
\label{subsec:Model}
A feature extraction model is modified from Resnet-50 to support one channel instead of three channels and use SELU as an activation function. The other parts are similar to the original Resnet-50 \cite{ResNet} composed of five stages of convolution blocks and identical blocks including a stack of convolutional blocks, batch normalization, and activation with skip connections in each block and output 128 features. The classification model is simply a linear layer with 10 neurons to classify 128 features from feature extraction into 10 classes.

\label{subsection:Training} We apply the same settings for all 1- to 3-day cases and all locations in both stages 1 and 2 to be consistent in all domain configurations. Every convolution layer applies $1\mathrm{e}{-4}$ of $l^2$ kernel regularizer and trains for 10,000 epochs. The training optimizer is ADAM with an initial learning rate starting at $1\mathrm{e}{-3}$ and ending at $1\mathrm{e}{-5}$, applying the polynomial learning rate decay scheduler with learning rate restart. We use a batch size of 64. The confidence threshold of the consistency pseudo-label is 0.97.  Additive Margin Softmax Loss 
applies $l^2$-normalization to both input and weights to the fully connected layer head without bias, with $s=10$ and $m=0.2$. All losses in both stages 1 and 2 are applied with equal weights except centroid alignment loss with $\alpha$ = 0.05 to gradually move centroids of the classes without breaking the current.

\subsection{Evaluation Results} We evaluate the model without any domain adaptation (supervised learning) as a baseline comparing with the five representative domain adaptation methods including GAN \cite{confusionLoss}, GRL \cite{GRL}, ADDA \cite{ADDA}, CDAN \cite{CDAN}, and FixMatch \cite{FixMatch} along with our method, GaitSADA, on the four different locations (laboratory, conference room, server room, and office) to explore the spatial domain adaptation using 1-3 days of data for training and the rest of data for testing. Meanwhile, We also study naive supervised learning without domain adaptation to understand the domain discrepancy issue in low data regime, temporal case, and spatial case.

As summarized in Table~\ref{tab:experiment}, our GaitSADA method outperforms the other five representative domain adaptation methods on all spatial domain adaptation (i.e., different locations) and temporal domain adaptation (i.e., different days on laboratory location). Without applying domain adaptation (i.e., supervised training in Table~\ref{tab:experiment}), the performance of 3-day cases is 79.96\% on average for all locations and temporal case, then drop to 65.61\% and 47.85\% in 2-day and 1-day data which apparently shows that the amount of data helps mitigate the domain shift issues in both spatial and temporal cases.



GAN, GRL, and ADDA yield similar results and help alleviate the domain drift issue by boosting accuracy from the supervised method on average in all 3-day cases. However, we observe the accuracies drop compared to the supervised learning method in 1-day case for all these three methods, e.g., GAN and ADDA drop 0.46\% and 0.79\% in office location, GRL drops 1.08\% for the temporal case due to the intricate structure of minimax game between domain discriminator and feature extraction. The domain discriminator is unable to entirely minimize the domain shift while training on fewer data since it does not learn all feature distinctions of the source and target domains.

CDAN has better results comparing to GAN, GRL, and ADDA, 56.12\%, 76.04\%, and 92.35\% on average of 1-day, 2-day, and 3-day respectively due to its ability to learn class-wise feature distribution cross-covariance dependency on class prediction and its feature, therefore, helping understanding the class-conditional domain drift. However, the accuracy drop happens in 1-day temporal case which CDAN accuracy is 73.67\% while supervised learning method is 76.05\%. The limited amount of training data is still a problem for CDAN as it requires more data to understand the distribution of source and target.

FixMatch performs best comparing to the other representative methods even though its original purpose is for semi-supervised learning tasks, not unsupervised domain adaptation. Because FixMatch synthesizes data utilizing pseudo-label from target unlabeled data to train the model, it virtually trains on more data. We observe that the advantage of FixMatch presents more than the other domain adaptation methods in low data regime as in the temporal 1-day case, FixMatch gives 18.27\% performance boost from supervised learning method while the other methods struggle with adapting for the small dataset. 


The average accuracies for 1-day, 2-day, and 3-day cases of GaitSADA are 79.66\%, 91.80\%, and 97.57\% respectively which increase accuracies of vanilla supervised learning method by a large margin and outperform other baselines. In particular, our results for the 1-day scenario clearly demonstrate a very strong accuracy improvement in the low data regime due to the simplicity of the aligning components without a complex discriminating structure that requires more training data (e.g., GAN, GRL, ADDA). Specifically, GaitSADA provides performance boosts  by  31.82\%, 26.19\%, and 17.61\% in terms of average accuracy of 1-day, 2-day, and 3-day cases respectively, compared with the vanilla supervised learning method, and by 15.41\%, 5.83\%, and 2.63\%, compared with FixMatch, which is the best baseline.



\subsection{Visualization}
\begin{figure}[t] 
\centerline{\includegraphics[width=.5\textwidth]{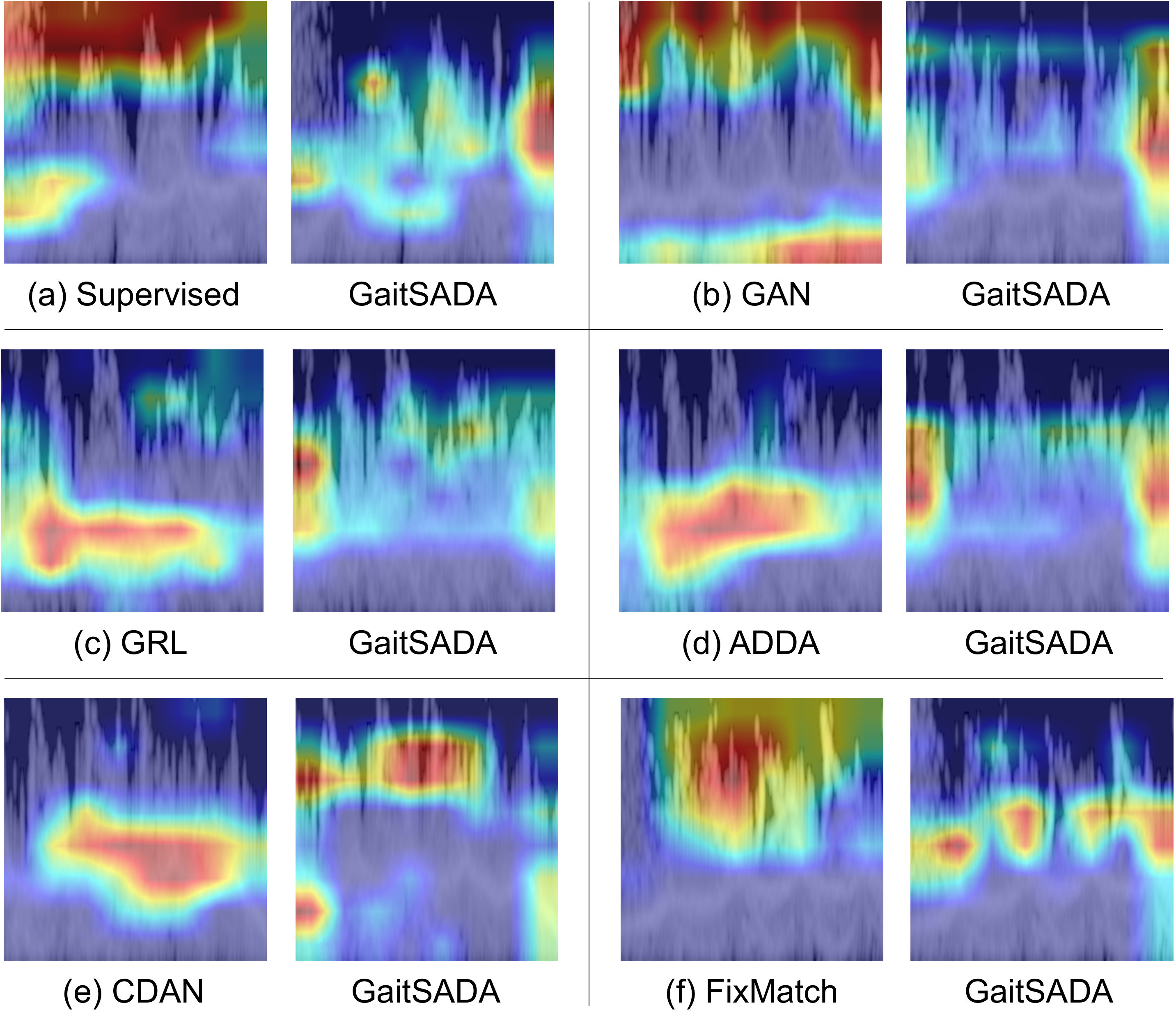}}

\caption{Grad-CAM \cite{grad-CAM} visualizations of the wrong prediction of other methods (a)-(f) comparing to our correct prediction of the same data. The red color area represents the more important weight contributing to the prediction. }
\label{fig:gradcam-selected}
\end{figure}

Grad-CAM \cite{grad-CAM} visualizes the weighted feature maps to create a heatmap that highlights the most important regions to make a prediction. Fig.~\ref{fig:gradcam-selected} illustrates diverse Grad-CAM saliency heatmaps for each algorithm. The other methods' wrong predictions tend to focus only on a specific part of the spectrogram data, high-frequency or low-frequency part. The high-frequency part represents the fast-moving limbs i.e., arms and legs, indicating the importance of this region for gait recognition. However, (a) Supervised, (b) GAN, and (f) FixMatch show the models tend to confuse high-frequency and the white noise in the background. Concentrating on this background noise area and ignoring the other parts lead to misprediction. (c) GRL, (d) ADDA, and (e) CDAN tend to focus only on the low-frequency part, the torso region, which may contain less unique information compared to other parts of the body. We visualize saliency of GaitSADA on the same data that the other methods predict incorrectly. In contrast to the other methods, GaitSADA looks broadly at multiple parts of the body without focusing too much on specific parts which allow our method to effectively use more unique information from different parts of the body for gait recognition and thus easily adapt to domain shift.

\subsection{Ablation Study}

\begin{table*}[h]
\caption{Ablation experiments of 2-day data on laboratory location as source data while adapting to different days
(i.e., temporal) and 3 different locations, (i.e., server, conference, and office) using multiple combinations of Additive Margin, Contrastive learning, Consistency Training and Centroid Alignment. The results report in \% of accuracy.}
\centering
\begin{adjustbox}{width=0.8\linewidth,center}
\begin{tabular}{c|l|cccc|c} 
\hline
\multicolumn{1}{l|}{\textbf{Stage}}                                                                 & \multicolumn{1}{c|}{\textbf{Components}}                                                                                                                             & \textbf{Temporal} & \textbf{Server} & \textbf{Conference} & \textbf{Office} & \textbf{Average}  \\ 
\hline
\multirow{4}{*}{\begin{tabular}[c]{@{}c@{}}\textbf{1 stage}\\\textbf{(train jointly)}\end{tabular}} & 1. supervised                                                                                                                                                        & 88.46             & 53.05           & 71                  & 49.92           & 65.61             \\
                                                                                                    & 2. supervised + AM                                                                                                                                                   & 91.8              & 73.72           & 82.8                & 67.07           & 78.85             \\
                                                                                                    & 3. supervised + AM + contrastive learning                                                                                                                                 & 94..21            & \textbf{80.38}  & 89.37               & 70.87           & 83.71             \\
                                                                                                    & 4. supervised + AM + consistency + centroid                                                                                                                          & \textbf{99.19}    & 52.15           & 98.2                & 46.55           & 74.02             \\ 
\hline
\multirow{2}{*}{\textbf{2 stages}}                                                                  & \begin{tabular}[c]{@{}l@{}}5. \textbf{(1st stage)} supervised + AM + contrastive learning\\~ ~\textbf{ (2nd stage)} supervised + AM + consistency\end{tabular}             & 94.68             & 78.58           & 90.3                & 72.77           & 84.08             \\ 
\cline{2-7}
                                                                                                    & \begin{tabular}[c]{@{}l@{}}6. \textbf{(1st stage) }supervised + AM + contrastive learning \\~ ~ \textbf{(2nd stage)} supervised + AM + consistency + centroid\end{tabular} & 97.63             & 79.88           & \textbf{99.10}      & \textbf{90.59}  & \textbf{91.80}    \\
\hline
\end{tabular}
\end{adjustbox}
\label{tab:ablation}
\end{table*}

We experiment with 2-day data on laboratory location as source data while adapting to different days (i.e., temporal) and 3 different locations, (i.e., server, conference, and office) in the same setting as the experiments in Table~\ref{tab:experiment} and examine the behavior of each component contributing to the final results in both jointly training and 2-stage training setting. We study six different combinations of components as shown in Table~\ref{tab:ablation}, i.e., supervised learning, additive margin softmax loss, contrastive learning, consistency training, and centroid alignment. In particular, The AM component alone improves the performance by a large margin. In fact, this component already outperforms most of the other adversarial-based methods as AM loss increases inter-class separability and intra-class compactness without training multiple models (i.e. feature extractor and domain discriminator) as  adversarial-based methods did, which require more data in order to learn the feature distribution. Contrastive learning further improves by 4.86\% on average as the model learns noise invariance on the target data. Moreover, it can be observed that two-stage approach outperforms ($84\%$ and $91.8\%$) the single-stage approach (65.61\%, 78.85\%, 83.71\%, and 74.02\%), which indicates the consistency-based fine-tuning at the second training stage plays an active role. In addition, it shows that by combining centroid alignment and consistency training can further improve the recognition accuracy from $84\%$ to $91.8\%$.

\section{Conclusion}
In this paper, we designed a system for gait recognition using a commercial mmWave radar, conducted multiple times and environments to study the domain discrepancy in various cases, and proposed a novel domain adaptation method designed explicitly for mmWave signals. We experiment with five regular domain adaptation methods to close the domain gaps. The experiments show the limitations of adversarial-based domain adaptation methods which require more data in order to learn the data distribution due to the complicated minimax game of the feature extractor and domain discriminator model. On the other hand, GaitSADA directly minimizes intra-class variation while maximizing inter-class deviation by utilizing self- and semi-supervised to learn target unlabelled data. Our proposed method outperforms standard domain adaptation methods by a large margin, particularly in the low data regime. Visualization and ablation studies further validate the effectiveness of GaitSADA.

\bibliographystyle{IEEEtran}
\bibliography{ref/unsupervised_domain_adaptation, ref/semi_supervised, ref/self_supervised, ref/gait_recog, ref/radio-based, ref/old_ref, ref/other}

\end{document}